\documentclass[10pt,twocolumn,letterpaper]{article}

\usepackage{cvpr}              

\usepackage{graphicx}
\usepackage{amsmath}
\usepackage{amssymb}
\usepackage{booktabs}
\usepackage{multirow}

\usepackage[pagebackref,breaklinks,colorlinks]{hyperref}

\usepackage[capitalize]{cleveref}
\crefname{section}{Sec.}{Secs.}
\Crefname{section}{Section}{Sections}
\Crefname{table}{Table}{Tables}
\crefname{table}{Tab.}{Tabs.}

\def\cvprPaperID{*****} 
\def\confName{CVPR}
\def\confYear{2022}

\begin{document}

\title{Person Re-Identification }

\author{Mustafa E. Chasmai\\
Indian Institute of Technology Delhi\\
{\tt\small cs1190341@iitd.ac.in}
\and
Tamajit Banerjee\\
Indian Institute of Technology Delhi\\
{\tt\small cs1190408@iitd.ac.in}
}
\maketitle

\begin{abstract}
   Person Re-Identification (Re-ID) is an important problem in computer vision-based surveillance applications, in which one aims to identify a person across different surveillance photographs taken from different cameras having varying orientations and field of views. Due to the increasing demand for intelligent video surveillance, Re-ID has gained significant interest in the computer vision community. 
   In this work, we experiment on some existing Re-ID methods that obtain state of the art performance in some open benchmarks. We qualitatively and quantitaively analyse their performance on a provided dataset, and then propose methods to improve the results. 
\end{abstract}

\section{Introduction}
\label{sec:intro}

Intelligent video surveillance has been gaining a lot of attention recently. Along with detection and tracking, re-identification is one of the major tasks needed to get a complete surveillance system, and has its own set of challenges. Person Re-ID aims to find the person of a query image in a different set of gallery images that may contain the same person. This becomes particularly challenging when the gallery  has images from a wide variety of orientations, occlusions, lighting conditions, and backgrounds.

Many past works \cite{he2021transreid, sharma2021person, wojke2018deep, zhang2017alignedreid, luo2019alignedreid++} belong to the distance  metric  learning  paradigm, where the problem is reduced to learning a suitable metric that can provide a partitioning of images having the same and different persons. Equivalently, a suitable feature representation is to be learnt such that the latent vectors corresponding to the same person should have a large similarity, while those corresponding to different persons should have lower similarity. The similarity is determined by selecting an appropriate similarity metric. Various metrics of similarity like the norm and the cosine similarity have been used in literature. The Re-ID problem is different from person classification because the persons are not restricted to a fixed set of classes, and the model is expected to be robust to queries for new persons it had not encountered during training.

\begin{figure}
    \centering
    \includegraphics[width=0.1\linewidth]{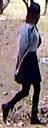}
    \includegraphics[width=0.1\linewidth]{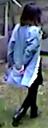}
    \includegraphics[width=0.1\linewidth]{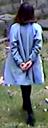}
    \includegraphics[width=0.1\linewidth]{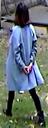}
    \includegraphics[width=0.1\linewidth]{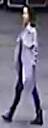}
    \includegraphics[width=0.1\linewidth]{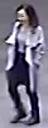}
    \includegraphics[width=0.1\linewidth]{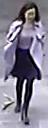}
    \includegraphics[width=0.1\linewidth]{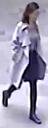}
    
    \includegraphics[width=0.1\linewidth]{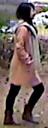}
    \includegraphics[width=0.1\linewidth]{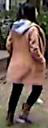}
    \includegraphics[width=0.1\linewidth]{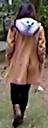}
    \includegraphics[width=0.1\linewidth]{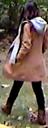}
    \includegraphics[width=0.1\linewidth]{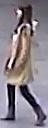}
    \includegraphics[width=0.1\linewidth]{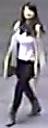}
    \includegraphics[width=0.1\linewidth]{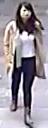}
    \includegraphics[width=0.1\linewidth]{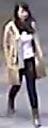}
    
    \includegraphics[width=0.1\linewidth]{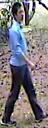}
    \includegraphics[width=0.1\linewidth]{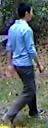}
    \includegraphics[width=0.1\linewidth]{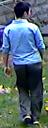}
    \includegraphics[width=0.1\linewidth]{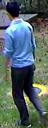}
    \includegraphics[width=0.1\linewidth]{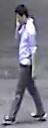}
    \includegraphics[width=0.1\linewidth]{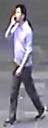}
    \includegraphics[width=0.1\linewidth]{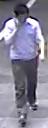}
    \includegraphics[width=0.1\linewidth]{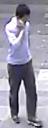}

    \caption{Few sample images from the ReID780 training dataset. Each row corresponds to different images for a single person. The images have varying orientation and poses of the concerned person, along with different lighting conditions and backgrounds.}
    \label{fig:data}
\end{figure}

There have been many different methods proposed to solve Re-ID. From hand crafted features to deep CNN based models to the recent transformer based models, the state of the art has improved considerably. These methods are benchmarked on various openly available datasets like Market-1501 \cite{zheng2015scalable}, CUHK03 \cite{li2014deepreid} and MSMT17 \cite{wei2018person}. We used the dataset provided by the instructors of COL780 , IIT Delhi for all the experimentation and analysis in this work. We qualitatively analyse the failure cases of the baseline and propose novel design and methodological changes to improve the performance further.

\begin{figure*}
    \centering
    \includegraphics[width=0.9\linewidth]{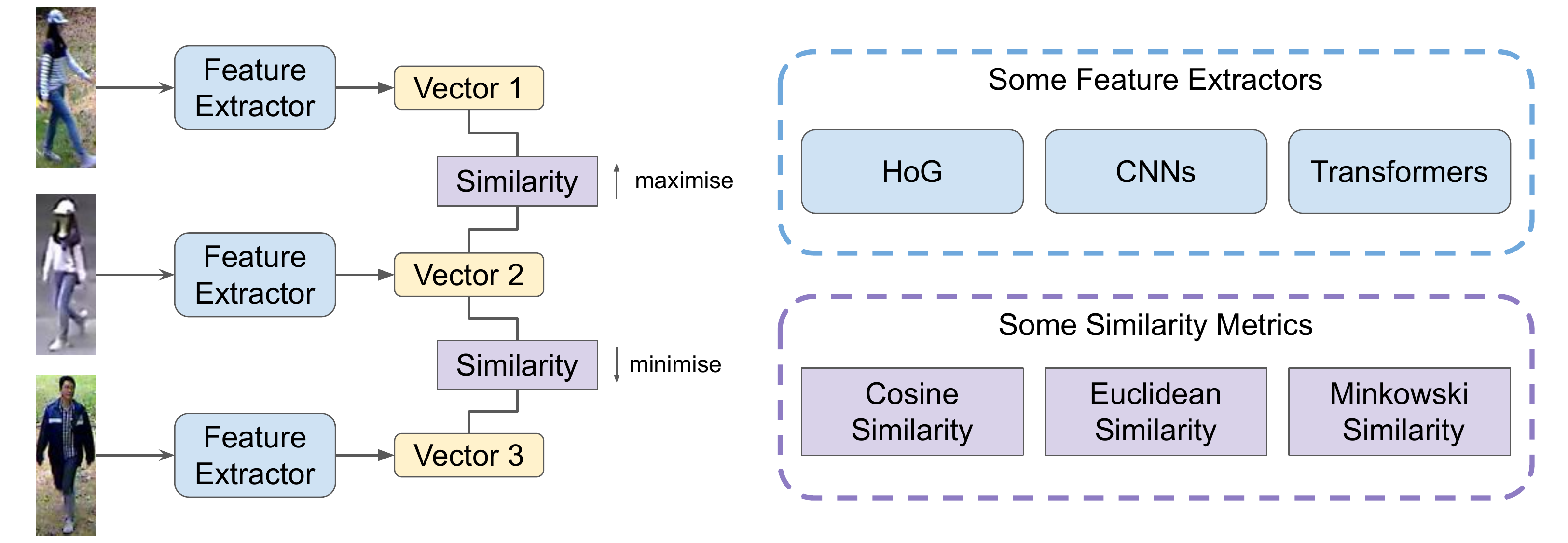}
    \caption{A rough outline of a generic architecture of a metric learning based method. The task is to learn a feature representation that maximises the similarity metric of images of the same person, and minimises the same for images of different persons.}
    \label{fig:generic}
\end{figure*}

\section{Related Work}
\label{sec:related}

There have been many methods proposed for Re-ID. Most early research in the field involved exploring new hand crafted features that could better represent a person's identity across varying viewpoints, lighting conditions and backgrounds. Some works \cite{gray2008viewpoint, farenzena2010person, liao2015person} focused on combining intuitive local features via ensembling or accumulation or maximal occurences, while others \cite{yang2014salient, matsukawa2016hierarchical} devised new feature representations. 

With the advent of deep learning, most methods shifted to CNNs for richer representations that could be learnt directly. Deep metric learning methods transform raw images into embedding features, then compute the feature similarities. The focus shifted from designing good features to designing suitable loss functions that could train the CNN based feature extractors (e.g. ResNet \cite{he2016deep}). The ID loss \cite{zheng2017discriminatively} and triplet loss \cite{liu2017end} are most widely used in deep ReID, with some methods \cite{luo2019bag, sun2020circle} using a combination of the two. The triplet loss uses three inputs, two of the same person, while one of a different one. It is designed so as to minimise the distance between the positive pair (two images of the same person), and to maximise the distance between the negative pair (different persons). There has also been some work in improving the strategy of selecting suitable positive and negative pairs. Ideas very similar to the triplet loss can be found in Siamese networks \cite{koch2015siamese}, Contrastive Learning \cite{chen2020simple} and GANs \cite{goodfellow2014generative}.

Many other methods have worked on improving the underlying architecture. AlignedReID \cite{zhang2017alignedreid, luo2019alignedreid++} extracts a global feature which is jointly learnt with local features by aligning them together. Aligning local features by pose estimation \cite{zheng2015scalable} has also been used in the past. Wojke et al. \cite{wojke2018deep} used a conventional softmax classification regime, and used the pre-softmax embeddings with small re-parametrisation to optimise the cosine similarity metric. 

The Transformer model \cite{vaswani2017attention}, proposed originally to handle sequential data in the field of natural language processing (NLP), has been gaining a lot of attention in the computer vision field. The Vision Transformer \cite{dosovitskiy2021an} uses the transformer architecture for image classification. The transformer has since been used for wide ranging tasks like segmentation \cite{wang2021end} and pose estimation \cite{stoffl2021end}, and some methods have used it for Re-ID as well. TransReID \cite{he2021transreid} and LA Transformers \cite{sharma2021person} are two sunch methods. There are many other methods like Centroids ReID \cite{wieczorek2021unreasonable} that also showed promise on existing datasets.

\section{Baselines}
\label{sec:baselines}

We experiment and explore 4 existing methods in person ReID, namely AlignedReID \cite{zhang2017alignedreid}, LA Transformers \cite{sharma2021person}, Centroid ReID \cite{wieczorek2021unreasonable} and TransReID \cite{he2021transreid}. We also ran Deep Cosine Metric \cite{he2016deep} on the dataset, but did not experiment on it further because of low accuracy and tensorflow - pytorch incompatibilities. The results of these baseline methods on the provided dataset can be found in Table~\ref{tab:baselines}.


\begin{table}[h]
    \centering
    \begin{tabular}{cccc}
    \toprule
        Method & mAP & CMC @R1 & CMC @R5\\
    \midrule
        AlignedReID & 98.5 & 100 & 100 \\
        LA-Transformer & 95.1 & 100 & 100\\
        Centroids-ReID & 68.4 & 82.1 & 82.1\\
        TransReID & 64.3 & 71.4 & 82.1\\
        Deep Cosine & 35 & - & - \\
    \bottomrule
    \end{tabular}
    \caption{Performance of some existing methods on the dataset}
    \label{tab:baselines}
\end{table}

We have used these codebases provided by the authors of the respective methods: \href{https://github.com/michuanhaohao/AlignedReID}{AlignedReID}, \href{https://github.com/SiddhantKapil/LA-Transformer}{LA Transformer}, \href{https://github.com/mikwieczorek/centroids-reid}{Centroids ReID}, \href{https://github.com/damo-cv/TransReID}{Trans ReID} and \href{https://github.com/nwojke/cosine_metric_learning}{Deep Cosine Metric}.

\begin{figure*}
    \centering
    \includegraphics[width=\textwidth]{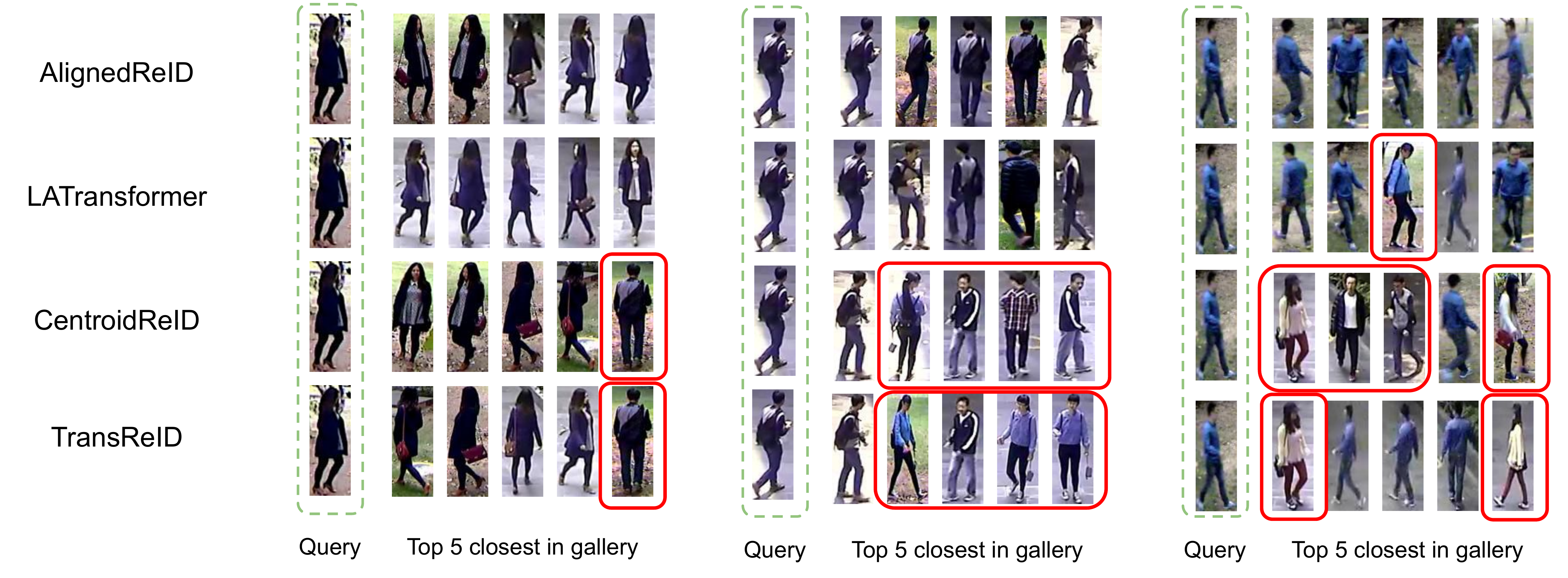}
    \caption{The top 5 closest images in the gallery for a few querry images with the explored baseline methods}
    \label{fig:preds}
\end{figure*}

\section{Analysis of Baselines}

We ran multiple experiments on the baselines to identify which is the best performing. We report these experimental observations, along with our interpretations and conclusions from them. 

The top 5 predictions based on similarity for a few query images can be seen in Fig~\ref{fig:preds}. It can be observed that AlignedReID and LA Transformer methods perform comparatively better than the rest. One interesting observation is that for the last query image, all of the models predicted the same wrong image, which was even the closest for three of them. We interpret this as an indication that there is some semantic similarity in them even though they look completely different to human eyes. Other than this corner case, all of the models seem to perform quite good, with even the wrong predictions being visually similar to the query image. These examples of visual similarity are especially difficult to deal with.

\begin{table}[h]
    \centering
    \begin{tabular}{cccc}
    \toprule
        Method & mAP & CMC @R1 & CMC @R5\\
    \midrule
        AlignedReID & 69.8 & 78.6 & 85.7 \\
        LA-Transformer & 75.2 & 89.3 & 82.1\\
        Centroids-ReID & 64.8 & 75.0 & 85.7\\
    \bottomrule
    \end{tabular}
    \caption{Performance of some existing methods on the dataset, trained on the original data and tested with the background masked}
    \label{tab:back_masks}
\end{table}

We wanted to explore the sensitivity of the models with the background information. Ideally, we expect a person re-identification model to be background agnostic, with no change in the predictions if the background is removed completely. We run the trained models for the baselines on images with their backgrounds masked out using a semantic segmentation model, and observe that in all of them, the performance drops. We interpret this as meaning that all of the models need some background context to be able to perform well, and in the absence of the background, they are not able to do so. From this, we conclude that the methods are somewhat overfitting on the background information.

\begin{figure}[h!]
    \centering
    \includegraphics[width=0.8\linewidth]{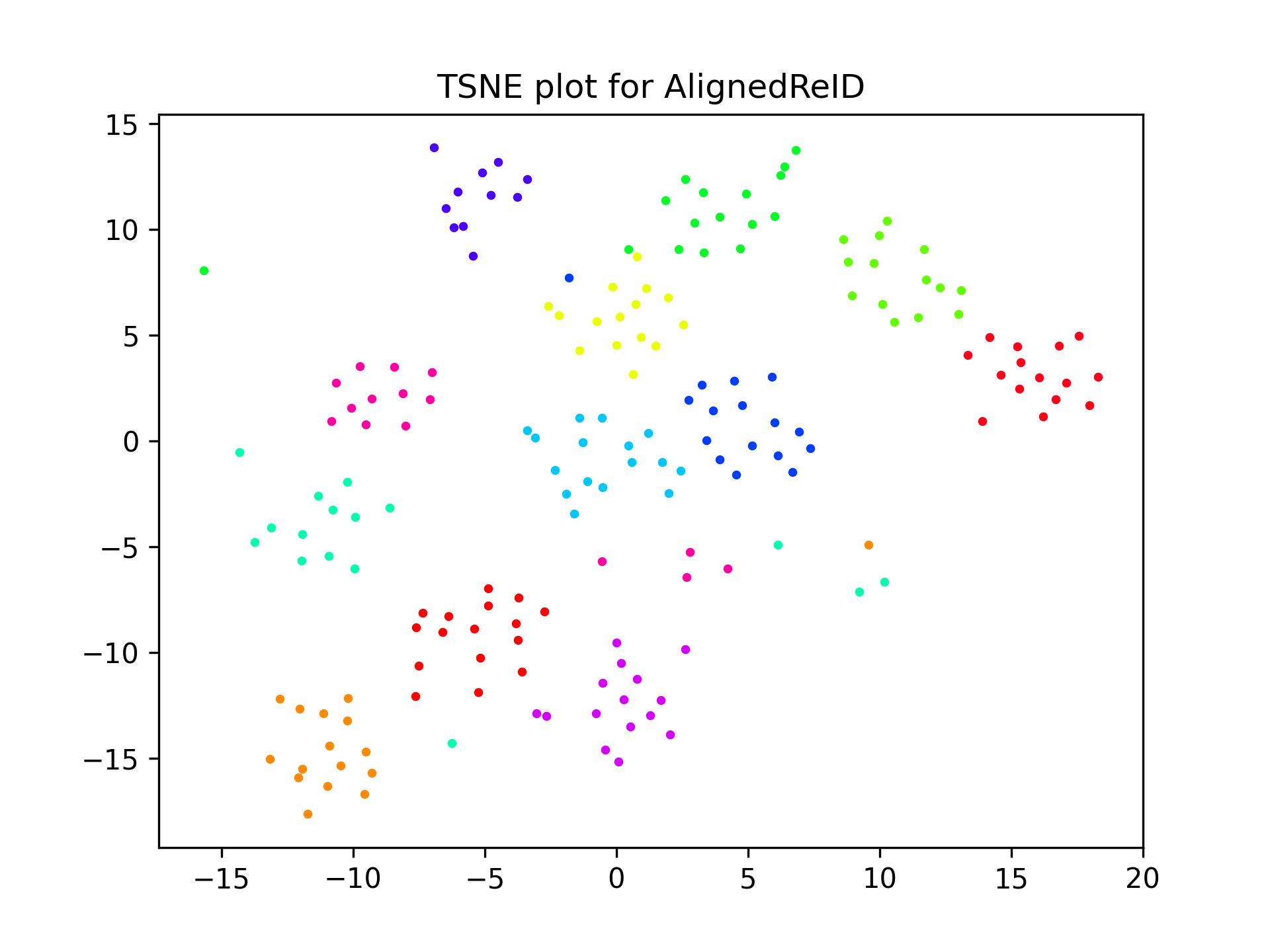}
    \includegraphics[width=0.8\linewidth]{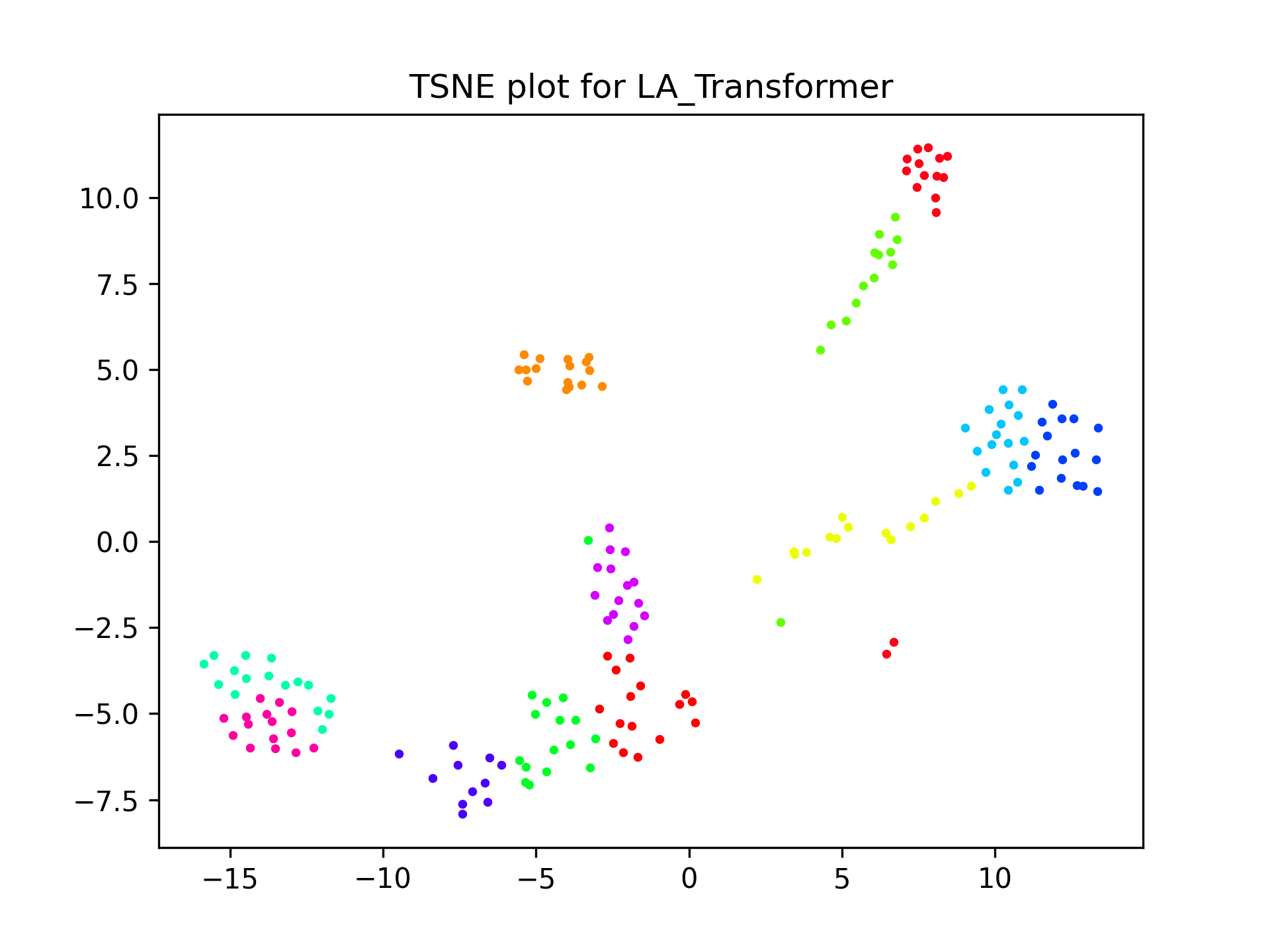}
    \caption{t-SNE plots AlignedReID (top) \& LATrans (down) }
    \label{fig:tsne}
\end{figure}

Finally, we wanted to explore how the learned features are distributed. We expect that the features of the same persons would be close together, while that of the different persons would be far from each other. Thus, we expect the formation of cluster-like structures in the feature space. We explore this for the case of the gallery images of the validation set, and plot t-SNE graphs for our different models. The plots can be seen in Fig~\ref{fig:tsne}. The t-SNE plots for LA Transformers seem to be slightly more clustered that the same for AlignedReID, and we conclude from this that LA Transformers learns more robust features.

\section{Proposed Methodologies}
\label{sec:methodology}
\subsection{Mask Guidance}

As we had analysed, removing the background pixels via masking leads to a significant drop in the performance of all the models. We wanted to make a methods that is invariant to the absence of backgrounds and thus, experiment with different ways of including the binary segmentation masks in the training process as well. 

Deriving inspiration from \cite{song2018mask}, we use the binary mask primarily for 2 reasons. It can improve the robustness of ReID models under various background conditions by removing background clutters in the pixel-level. Along with this, the mask contains body shape information which can be regarded as the important feature and also it is robust to illumination, color of clothes and thus is an important feature.

The most straightforward way to utilize the binary body mask is to directly mask the background in the images. With the binary mask, the masked image only contains the body region which is expected to perform better than using the whole image. However, in our experiments, we find the performance of masked images is even slightly worse compared with the one using the original images. This result means that directly removing the background with binary mask in a ‘hard’ manner is not a good choice, which may affect the structured information and smoothness of an image. In addition, the wrongly segmented masks may contain lots of backgrounds or lose some important body parts which will greatly impact the performance. In this case, removing the backgrounds in the feature-level may be a better solution.

We initially used only binary mask for predictions. Even though the mask did not contain colour information and only had body shapes. It gave an mAP of 29.2 which was better than random selection which showed that the body shape is important.

Our proposed solution is to send the mask along with the RGB image to learn the weights for feature. We add an extra convolutional layer at the beginning of the model. The main purpose of this layer is too merge the RGB image and the binary mask. Thus, with this, we have a model that takes as input the RGB image and binary mask, merges the two into a single feature in the first layer and then passes this through the remaining model same as used earlier. 

\subsection{Feature position invariance}

In Aligned ReId , the dataset on which they have worked does not have accurate bounding boxes and to counter that they have used the shortest distance algorithm through dynamic programming to find the distances between local feature vectors, which finally gets translated to the losses. 

On the other hand, our dataset is more organised, that is, each bounding box is nearly perfect. As a result, the local feature vectors should corresponding to each other better. So, the complex DP algorithm can be replaced by a simple one to one correspondence. This both reduced the time complexity and improves the performance on our dataset.

\subsection{Self Ensembling}

One observation we made during training the methods was that the validation performance metrics would fluctuate a lot as the model is trained. This was particularly prevalant in LA Transformer. To improve upon this short coming, we proposed the use of self ensembling. Self Ensembling is a powerful and easy to implement idea, as used in the Mean Teacher models \cite{yu2019uncertainty} for contrastive learning. 

The principle idea behind the mean teacher model is that of ensembling multiple weak models to obtain a much stronger model. The weak learners for self ensembling are the same model, at different stages of the training, particularly after training different batches. Self ensembling use a non-trainable clone of the actual model, and updates its parameters as the exponentially moving average of the training model. In literature, this exponentially moving averaged model is called the mean teacher model, and the training model as the student model. Finally, they use a consistency loss between the teacher and student such that the student learns the same features as the teacher. The teacher with the average weights is the ensemble of the models in the last few training steps, and is the main driving force for this self ensembling approach.

We observed that upon adding the self ensembling mean teacher approach, and using the consistency loss, the training of LA Transformer was comparatively much more smoother, and also the performance was better. Details can be found in the experimental results later.

\begin{figure*}
    \centering
    \includegraphics[width=\textwidth]{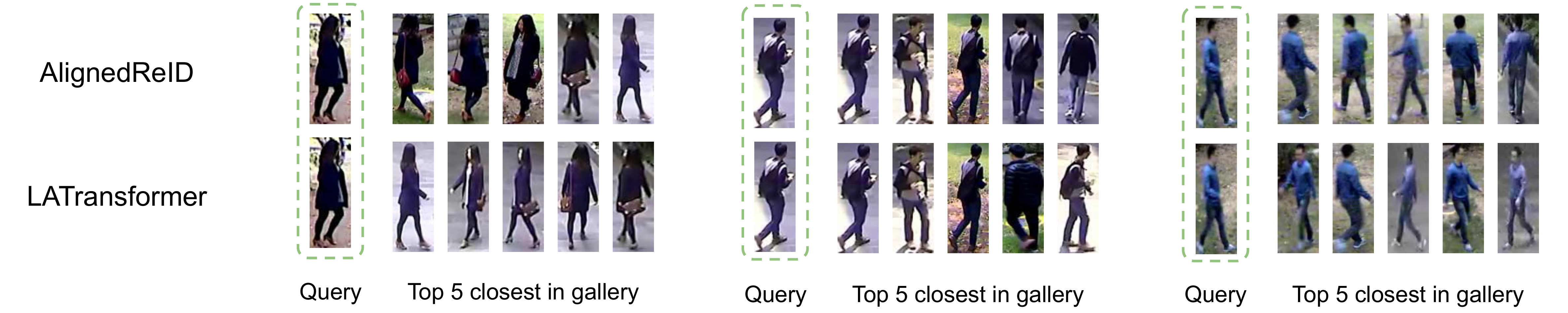}
    \caption{Predictions on the two improved models}
    \label{fig:imp_pred}
\end{figure*}

\begin{figure*}
    \centering
    \includegraphics[width=0.37\textwidth]{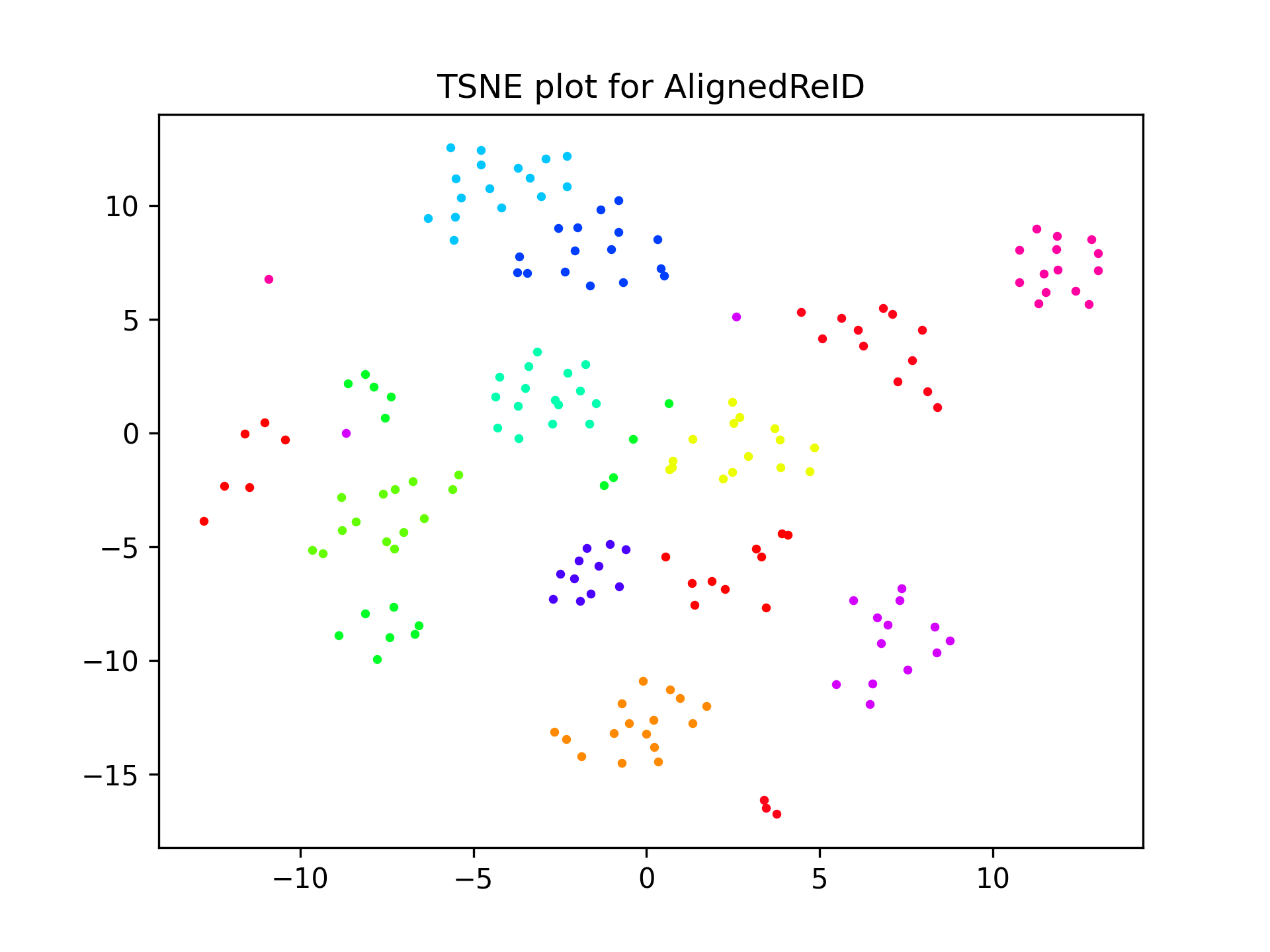}
    \includegraphics[width=0.37\textwidth]{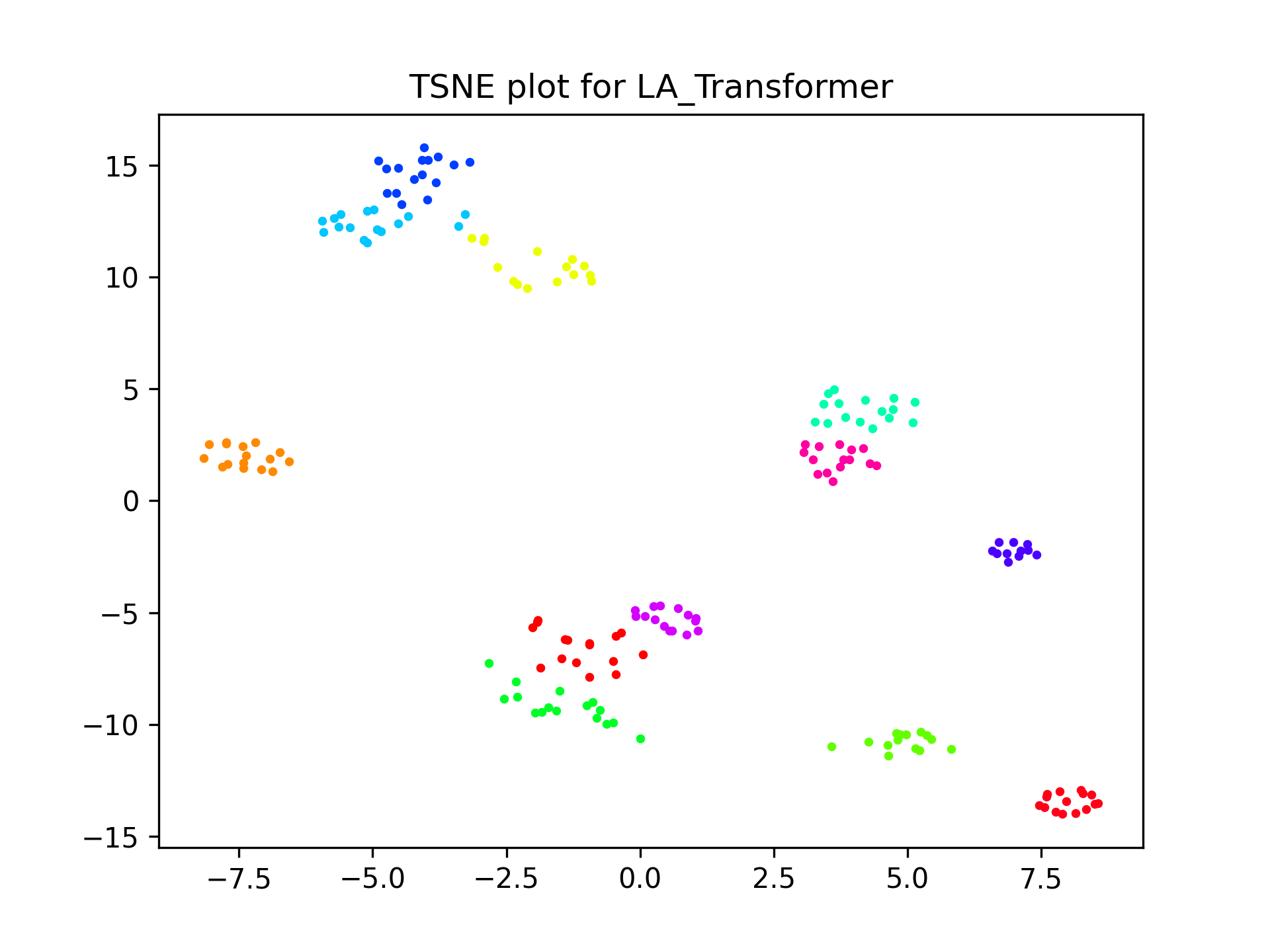}
    \caption{t-SNE plots for improved AlignedReID and LA Transformer}
    \label{fig:imp_tsne}
\end{figure*}

\subsection{Triplet Loss and Hard Example Mining}

LA Transformer uses only the cross entropy loss for training. We added triplet loss along with it, and expected a slightly better performance.

The basic idea is that the distance between the positive pair should be smaller than the negative pair. For calculating the loss, a triplet is formed, containing an anchor sample, a positive sample having the same person, and a negative sample from a different person. The distance between the anchor and positive sample is to be minimised while the same between the anchor and negative sample is to be maximised. This is very similar to the contrastive loss commonly used in self supervised learning. 

The performance of the triplet loss depends on the choice of positive and negative samples for given anchors. One widely used approach is Hard example mining. In this approach, the sample of the same person that is farthest and the sample of another class that is closest to the anchor is chosen. These samples are expected to be hardest to classify correctly, and thus using these as a triplet should guide the model towards learning better.

\subsection{Camera Contextual Learning}

The dataset provided to us had a more structured distribution compared to most existing datasets like Market1501. In this dataset, each person has images from 8 different angles, while in most other datasets, the number of camera angles present is often not as rigorous and mostly random. Thus, we though of a novel approach to leverage this structure in the data.

We reason that the change in the feature vectors between camera orientation should be agnostic of the identity of the person, and thus, should be nearly the same for all persons in the gallery. We wanted to explicitly encode this prior in the model, and share weights for the camera orientation dependent features. Thus, keeping the number of cameras as 8 in the dataset, we initialise 8 new residual layers that take as input the complete features and output embeddings of the same dimension. Since the residual layers make it easy to learn a unity function, these extra layers orientation should also be easily able to learn the non-camera contextual features. 

The main point is that these 8 layers have the same shared weights for all persons, but are different for different camera angles. Thus, the prior for camera orientation is effectively introduced to the model. In effect, we expected the model to be able to learn features that are clustered around a ring like shape instead of the usual point clusters, where the ring radius roughly corresponds to the extra residuals learned by the added camera layers. Thus, instead of trying to cluster features of the same person to the same point, we effectively try to cluster them in a ring like shape for each person. The fact that we observe a similar ring like pattern for the LA Transformer model is validation of our reasoning. We were not able to experiment this idea thoroughly, and our initial experiments indicated that the performance is actually dropping.

\begin{table*}[ht!]
    \centering
    \begin{tabular}{ccccc}
    \toprule
        Baseline & Experiments & mAP & CMC @Rank 1 & CMC @Rank 5\\
    \midrule
        \multirow{5}{*}{AlignedReID} & Feature Position Invariance & 99.0 & 100 & 100 \\
         & Binary Masks & 29.2 & 30.1 & 52.4 \\
        & masked RGB images & 76.7 & 90.8  & 83.6 \\ 
         & RGB images + masks &  88.0 & 92.9 & 100\\
        & Camera Contextual Learning & 67.3 & 78.6 & 85.7 \\
        \midrule
        \multirow{3}{*}{La Transformer} & Triplet Loss & 94.3 & 100 & 100 \\
        & Self Ensembling & 96.4 & 100 & 100 \\
         & Triplet Loss + Self Ensembling & 98.4 & 100 & 100 \\
    \bottomrule
    \end{tabular}
    \caption{List of experiments performed and the metrics obtained in each}
    \label{tab:res}
\end{table*}

\begin{table*}[ht!]
    \centering
    \begin{tabular}{ccccc}
    \toprule
         Baseline & Improvement Method & mAP & CMC @Rank 1 & CMC @Rank 5\\
    \midrule
       AlignedReID & Feature Position Invariance & 99.0 & 100 & 100 \\
        La Transformer & Triplet Loss + Self Ensembling & 98.4 & 100 & 100 \\
    \bottomrule
    \end{tabular}
    \caption{Experiments that led to an improvement over the baaseline}
    \label{tab:res_imp}
\end{table*}

\section{Experimental Results and Analysis}
\label{sec:results}








We trained and tested all of our proposed methodologies on the provided dataset. The quantitaive results for the different experiments are tabulated in Table~\ref{tab:res}, while the best performing methods are summarised in Table~\ref{tab:res_imp}.

In addition to the quantitative results, we report a qualitative analysis of the methods. The visualisation of top 5 predictions for the two improved methods can be seen in Fig~\ref{fig:imp_pred}, while the t-SNE plots for the same can be seen in Fig~\ref{fig:imp_tsne}. Comparing these with similar visualisations of the baselines, we can see that the improvements are reasonable, and that we can expect similar performance boosts in the test dataset. The trained weights of our baseline and improved models can be found \href{https://drive.google.com/drive/u/0/folders/1J4zorTwkK3McNUiwBKU67xi8oCW_fobr}{here}. The training code can be found in the corresponding colab links: \href{https://colab.research.google.com/drive/1lyKRB1ksg5Kks2So7XUCaUZf6iZwgCf5#offline=true&sandboxMode=true}{LATransformer}, \href{https://colab.research.google.com/drive/1Xt0HzneUmxESpmnns_0xOJLq2e8e2cC-#offline=true&sandboxMode=true}{AlignedReID}, \href{https://colab.research.google.com/drive/18CR4WFyHrLS34XUMmulWAdQuIWWbqaHX#offline=true&sandboxMode=true}{Centroids-ReID} and \href{https://colab.research.google.com/drive/1wLQFLXTuoteNK_CC3xt1G_XRVFG1BXtO#offline=true&sandboxMode=true}{TransReID}.

        


\section{Conclusion}
\label{sec:conclusion}

Person Re-identification is a challenging and yet unsolved task in Computer Vision that has seed a rising popularity in recent. We thoroughly analyse few state of the art approaches in this domain on the dataset provided, and try to point out some poossible research gaps. Based on these analyses, we propose  novel approaches to bridge these gaps.

First, we experiment with mask guidance. Though the mask guidance method showed promise, we were not able to achieve good results with the method. We suspect that the model was not able to properly incorporate the semantic information contained in the binary mask into its training. We wanted to propagate this mask information into the deeper layers of our model, and expect this to improve the performance of our model by making it more robust to background variation. Next, we observed that while AlignedReID had some mechanisms for images where the bounding boxes are not perfectly aligned, these cases are not present in the provided dataset, and removing them led to some improvements in the method. 

We further explore other methods to improve the performance of the model. We use self-ensembling and triplet loss with hard example mining to improve the performance of LA-Transformer. Finally, we make architectural changes in LA-Transformer to introduce the domain specific prior of camera orientations. Although the performance does not improve in some of our proposed approaches, we believe that they are in the right direction, and have scope for improvement.


\begin{thebibliography}{10}\itemsep=-1pt

\bibitem{chen2020simple}
Ting Chen, Simon Kornblith, Mohammad Norouzi, and Geoffrey Hinton.
\newblock A simple framework for contrastive learning of visual
  representations.
\newblock In {\em International conference on machine learning}, pages
  1597--1607. PMLR, 2020.

\bibitem{dosovitskiy2021an}
Alexey Dosovitskiy, Lucas Beyer, and Alexander~Kolesnikov et al.
\newblock An image is worth 16x16 words: Transformers for image recognition at
  scale.
\newblock In {\em International Conference on Learning Representations}, 2021.

\bibitem{farenzena2010person}
Michela Farenzena, Loris Bazzani, Alessandro Perina, Vittorio Murino, and Marco
  Cristani.
\newblock Person re-identification by symmetry-driven accumulation of local
  features.
\newblock In {\em 2010 IEEE computer society conference on computer vision and
  pattern recognition}, pages 2360--2367. IEEE, 2010.

\bibitem{goodfellow2014generative}
Ian Goodfellow, Jean Pouget-Abadie, Mehdi Mirza, Bing Xu, David Warde-Farley,
  Sherjil Ozair, Aaron Courville, and Yoshua Bengio.
\newblock Generative adversarial nets.
\newblock {\em Advances in neural information processing systems}, 27, 2014.

\bibitem{gray2008viewpoint}
Douglas Gray and Hai Tao.
\newblock Viewpoint invariant pedestrian recognition with an ensemble of
  localized features.
\newblock In {\em European conference on computer vision}, pages 262--275.
  Springer, 2008.

\bibitem{he2016deep}
Kaiming He, Xiangyu Zhang, Shaoqing Ren, and Jian Sun.
\newblock Deep residual learning for image recognition.
\newblock In {\em Proceedings of the IEEE conference on computer vision and
  pattern recognition}, pages 770--778, 2016.

\bibitem{he2021transreid}
Shuting He, Hao Luo, Pichao Wang, Fan Wang, Hao Li, and Wei Jiang.
\newblock Transreid: Transformer-based object re-identification.
\newblock {\em arXiv preprint arXiv:2102.04378}, 2021.

\bibitem{koch2015siamese}
Gregory Koch, Richard Zemel, Ruslan Salakhutdinov, et~al.
\newblock Siamese neural networks for one-shot image recognition.
\newblock In {\em ICML deep learning workshop}, volume~2. Lille, 2015.

\bibitem{li2014deepreid}
Wei Li, Rui Zhao, Tong Xiao, and Xiaogang Wang.
\newblock Deepreid: Deep filter pairing neural network for person
  re-identification.
\newblock In {\em Proceedings of the IEEE conference on computer vision and
  pattern recognition}, pages 152--159, 2014.

\bibitem{liao2015person}
Shengcai Liao, Yang Hu, Xiangyu Zhu, and Stan~Z Li.
\newblock Person re-identification by local maximal occurrence representation
  and metric learning.
\newblock In {\em Proceedings of the IEEE conference on computer vision and
  pattern recognition}, pages 2197--2206, 2015.

\bibitem{liu2017end}
Hao Liu, Jiashi Feng, Meibin Qi, Jianguo Jiang, and Shuicheng Yan.
\newblock End-to-end comparative attention networks for person
  re-identification.
\newblock {\em IEEE Transactions on Image Processing}, 26(7):3492--3506, 2017.

\bibitem{luo2019bag}
Hao Luo, Youzhi Gu, Xingyu Liao, Shenqi Lai, and Wei Jiang.
\newblock Bag of tricks and a strong baseline for deep person
  re-identification.
\newblock In {\em Proceedings of the IEEE/CVF Conference on Computer Vision and
  Pattern Recognition Workshops}, pages 0--0, 2019.

\bibitem{luo2019alignedreid++}
Hao Luo, Wei Jiang, Xuan Zhang, Xing Fan, Jingjing Qian, and Chi Zhang.
\newblock Alignedreid++: Dynamically matching local information for person
  re-identification.
\newblock {\em Pattern Recognition}, 94:53--61, 2019.

\bibitem{matsukawa2016hierarchical}
Tetsu Matsukawa, Takahiro Okabe, Einoshin Suzuki, and Yoichi Sato.
\newblock Hierarchical gaussian descriptor for person re-identification.
\newblock In {\em Proceedings of the IEEE conference on computer vision and
  pattern recognition}, pages 1363--1372, 2016.

\bibitem{sharma2021person}
Charu Sharma, Siddhant~R Kapil, and David Chapman.
\newblock Person re-identification with a locally aware transformer.
\newblock {\em arXiv preprint arXiv:2106.03720}, 2021.

\bibitem{song2018mask}
Chunfeng Song, Yan Huang, Wanli Ouyang, and Liang Wang.
\newblock Mask-guided contrastive attention model for person re-identification.
\newblock In {\em Proceedings of the IEEE conference on computer vision and
  pattern recognition}, pages 1179--1188, 2018.

\bibitem{stoffl2021end}
Lucas Stoffl, Maxime Vidal, and Alexander Mathis.
\newblock End-to-end trainable multi-instance pose estimation with
  transformers.
\newblock {\em arXiv preprint arXiv:2103.12115}, 2021.

\bibitem{sun2020circle}
Yifan Sun, Changmao Cheng, Yuhan Zhang, Chi Zhang, Liang Zheng, Zhongdao Wang,
  and Yichen Wei.
\newblock Circle loss: A unified perspective of pair similarity optimization.
\newblock In {\em Proceedings of the IEEE/CVF Conference on Computer Vision and
  Pattern Recognition}, pages 6398--6407, 2020.

\bibitem{vaswani2017attention}
Ashish Vaswani, Noam Shazeer, Niki Parmar, Jakob Uszkoreit, Llion Jones,
  Aidan~N Gomez, {\L}ukasz Kaiser, and Illia Polosukhin.
\newblock Attention is all you need.
\newblock In {\em Advances in neural information processing systems}, pages
  5998--6008, 2017.

\bibitem{wang2021end}
Yuqing Wang, Zhaoliang Xu, Xinlong Wang, Chunhua Shen, Baoshan Cheng, Hao Shen,
  and Huaxia Xia.
\newblock End-to-end video instance segmentation with transformers.
\newblock In {\em Proceedings of the IEEE/CVF Conference on Computer Vision and
  Pattern Recognition}, pages 8741--8750, 2021.

\bibitem{wei2018person}
Longhui Wei, Shiliang Zhang, Wen Gao, and Qi Tian.
\newblock Person transfer gan to bridge domain gap for person
  re-identification.
\newblock In {\em Proceedings of the IEEE conference on computer vision and
  pattern recognition}, pages 79--88, 2018.

\bibitem{wieczorek2021unreasonable}
Mikolaj Wieczorek, Barbara Rychalska, and Jacek Dabrowski.
\newblock On the unreasonable effectiveness of centroids in image retrieval.
\newblock {\em arXiv preprint arXiv:2104.13643}, 2021.

\bibitem{wojke2018deep}
Nicolai Wojke and Alex Bewley.
\newblock Deep cosine metric learning for person re-identification.
\newblock In {\em 2018 IEEE winter conference on applications of computer
  vision (WACV)}, pages 748--756. IEEE, 2018.

\bibitem{yang2014salient}
Yang Yang, Jimei Yang, Junjie Yan, Shengcai Liao, Dong Yi, and Stan~Z Li.
\newblock Salient color names for person re-identification.
\newblock In {\em European conference on computer vision}, pages 536--551.
  Springer, 2014.

\bibitem{yu2019uncertainty}
Lequan Yu, Shujun Wang, Xiaomeng Li, Chi-Wing Fu, and Pheng-Ann Heng.
\newblock Uncertainty-aware self-ensembling model for semi-supervised 3d left
  atrium segmentation.
\newblock In {\em International Conference on Medical Image Computing and
  Computer-Assisted Intervention}, pages 605--613. Springer, 2019.

\bibitem{zhang2017alignedreid}
Xuan Zhang, Hao Luo, Xing Fan, Weilai Xiang, Yixiao Sun, Qiqi Xiao, Wei Jiang,
  Chi Zhang, and Jian Sun.
\newblock Alignedreid: Surpassing human-level performance in person
  re-identification.
\newblock {\em arXiv preprint arXiv:1711.08184}, 2017.

\bibitem{zheng2015scalable}
Liang Zheng, Liyue Shen, Lu Tian, Shengjin Wang, Jingdong Wang, and Qi Tian.
\newblock Scalable person re-identification: A benchmark.
\newblock In {\em Proceedings of the IEEE international conference on computer
  vision}, pages 1116--1124, 2015.

\bibitem{zheng2017discriminatively}
Zhedong Zheng, Liang Zheng, and Yi Yang.
\newblock A discriminatively learned cnn embedding for person reidentification.
\newblock {\em ACM Transactions on Multimedia Computing, Communications, and
  Applications (TOMM)}, 14(1):1--20, 2017.

\end{thebibliography}

\end{document}